\documentclass[10pt,journal]{IEEEtran}
\usepackage{url}
\usepackage{latexsym}
\usepackage[noend]{algpseudocode}
\usepackage{tabularx}
\usepackage{multirow}
\usepackage{amsmath}
\usepackage[normalem]{ulem}
\usepackage{mathrsfs}
\usepackage{amssymb}
\usepackage{amsthm}
\usepackage[dvips]{graphicx}
\usepackage{epsfig}

\newtheorem{theorem}{Theorem}
\newtheorem{definition}{Definition}
\newtheorem{lemma}[theorem]{Lemma}

\newcommand{\argmax}{\operatornamewithlimits{argmax}}

\def\bSig\pmb{\Sigma}

\begin{document}

\title{Exact Decoding on Latent Variable Conditional Models is NP-Hard}

\author{Xu Sun
\IEEEcompsocitemizethanks{
\IEEEcompsocthanksitem X. Sun is with Department of Computer Science, School of EECS, Peking University.
E-mail: xusun@pku.edu.cn
}
\thanks{}}

\IEEEcompsoctitleabstractindextext{%
\begin{abstract}
Latent variable conditional models, including the latent conditional random fields as a special case, are popular models for many natural language processing and vision processing tasks. The computational complexity of the exact decoding/inference in latent conditional random fields is unclear. In this paper, we try to clarify the computational complexity of the exact decoding. We analyze the complexity and demonstrate that it is an NP-hard problem even on a sequential labeling setting. Furthermore, we propose the latent-dynamic inference (LDI-Naive) method and its bounded version (LDI-Bounded), which are able to perform exact-inference or almost-exact-inference by using top-$n$ search and dynamic programming.
\end{abstract}

\begin{keywords}
Latent Variable Conditional Models, Computational Complexity Analysis, Exact Decoding.
\end{keywords}}

\maketitle

\IEEEdisplaynotcompsoctitleabstractindextext
\IEEEpeerreviewmaketitle

\section{Introduction}

Real-world problems may contain hidden
structures that are difficult to be captured by conventional
structured classification models without latent variables. For example, in the syntactic parsing task for natural language, the hidden structures can be refined grammars that are unobservable in the supervised training data \cite{PetrovK08}. In the gesture recognition task of the computational vision area, there are also hidden structures which are crucial for successful gesture recognition \cite{Wang06}. There are also plenty of hidden structure examples in other tasks among different areas \cite{QuattoniCD04,MorencyQ07,Sun08,sunACL09,Yu09,Wang09,Petrov10}

In such cases, models that exploit latent variables
are advantageous in learning \cite{QuattoniCD04,MorencyQ07,PetrovK08,Sun08,sunACL09,SunNAACL09,SunIJCAI09,SunT09,Petrov10}. As a representative structured classification model with latent variables,
the latent conditional random fields
(LCRFs) have become widely-used for performing a variety of tasks with
hidden structures, e.g., vision recognition
\cite{MorencyQ07}, and syntactic parsing
\cite{PetrovK08,Petrov10}.
For example, \cite{MorencyQ07} demonstrated that
LCRF models can learn latent structures of vision recognition problems
efficiently, and outperform several widely-used conventional models,
such as support vector machines (SVMs), conditional random fields (CRFs)
and hidden Markov models (HMMs). \cite{PetrovK08} and \cite{Petrov10} reported on a syntactic parsing task
that LCRF models can learn more accurate grammars than
models that use conventional techniques without latent variables.

Exact inference in the latent
conditional models is a remaining problem. In conventional models,
such as conditional random fields (CRFs), the optimal labeling can be efficiently obtained by
dynamic programming algorithms (for example, the Viterbi algorithm). Nevertheless, for latent conditional random fields, the inference is not straightforward, because of the
inclusion of latent variables. The computational complexity of inference in LCRFs, with a disjoint association among latent variables and a linear chain structure, is still unclear.
In this paper, we will show that such kind of inference
is actually NP-hard.
This is a critical limitation on the real-world applications of LCRFs. Most of the previous applications of LCRFs tried to make simplified approximations on the inference \cite{MatsuzakiM05,MorencyQ07,PetrovK08}, but the inference accuracy can be limited.

Although we will show that the inference in LCRFs is NP-hard, in real-world applications, we have an interesting observation on LCRF
models: they normally have a highly concentrated probability
distribution. The major probability is distributed on top-$n$ ranked
latent-labelings. In this paper, we try to make systematic analysis on this probability concentration phenomenon. We will show that probability concentration is reasonable and expected in latent conditional random fields.
Based on this analysis, we will propose a new inference algorithm: latent dynamic
inference (LDI), by systematically combining efficient top-$n$ search
with dynamic programming. The LDI is an exact inference
method, producing the most probable label sequence. In addition, for speeding up the inference, we
will also propose a bounded version of the LDI algorithm.

\section{Latent Conditional Random Fields}

\begin{figure}[t]
\begin{center}
\begin{tabular}{c}
    \includegraphics[width=0.8\hsize]{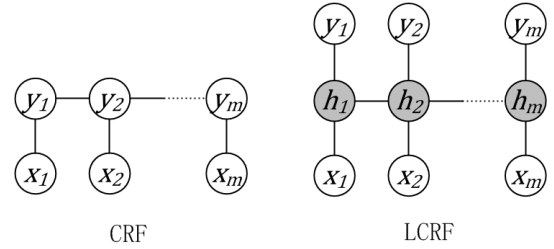}
\end{tabular}
\caption{Comparison between CRF models and LCRF models on the
training stage. $x$ represents the observation sequence, $y$
represents labels and $h$ represents the latent variables assigned
to the labels. Note that only the white circles are observed
variables. Also, only the links with the current observations are
depicted, but for both models, long range dependencies are possible.
}\label{fig.models}
\end{center}
\end{figure}

Given the training data, the task is to learn a mapping between a
sequence of observations $\pmb{x}=x_1,x_2,\ldots,x_m$ and a
sequence of labels $\pmb{y}=y_1,y_2,\ldots,y_m$. Each $y_j$ is a
class label for the $j$'th token of a word sequence, and is a member
of a set $\mathcal{Y}$ of possible class labels. For each sequence,
the model also assumes a sequence of latent variables
$\pmb{h}=h_1,h_2,\ldots,h_m$, which is unobservable in training
examples. See Figure~\ref{fig.models} for the comparsion between CRFs and LCRFs.

\begin{figure}[t]
\begin{center}
\begin{tabular}{c}
    \includegraphics[width=1\hsize]{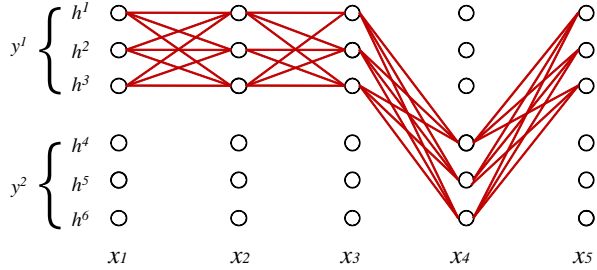}
\end{tabular}
\caption{In this example, the probability of the labeling ``$y^1,y^1,y^1,y^2,y^1$'' is summed over
all of its latent-labelings (represented by the red paths from position $x_1$ to $x_5$). The label $y^1$ has latent variables $h^1$ to $h^3$. The label $y^2$ has latent variables $h^4$ to $h^6$.
}\label{fig.disjoint}
\end{center}
\end{figure}

The LCRF model is defined as follows \cite{MorencyQ07}:
\begin{equation*} 
P(\pmb{y}|\pmb{x},\pmb{w})
\triangleq
\sum_{\pmb{h}}
P(\pmb{y}|\pmb{h},\pmb{x},\pmb{w})P(\pmb{h}|\pmb{x},\pmb{w}),
\end{equation*}
where $\pmb{w}$ represents the parameter vector of the
model.
To make the training efficient, a restriction is made for the model: for each label, the latent variables associated with it have no intersection with the latent variables from other labels \cite{MorencyQ07,PetrovK08}. This simplification is also a popular practice in other latent conditional models, including hidden-state conditional random fields (HCRF) \cite{QuattoniW08}. Each $h$ is a member in a set
$\mathcal{H}({y})$ of possible latent variables for the class label
$y$, and $\mathcal{H}(y^j) \cap \mathcal{H}(y^k) = \emptyset$ if $y^j \neq y^k$. $\mathcal{H}$ is defined as the set of all possible latent
variables; i.e., the union of all $\mathcal{H}({y})$ sets: $\mathcal{H} = \cup_{y \in \mathcal{Y}} \mathcal{H}(y)$. In other words, $h$ can have any value from $\mathcal{H}$, but $P(y|h)$ is zero except for only one of $y$ in $\mathcal{Y}$. The disjoint restriction indicates a discrete simplification of $P(\pmb{y}|\pmb{h},\pmb{x},\pmb{w})$:
\begin{equation*}
\begin{split}
&P(\pmb{y}|\pmb{h},\pmb{x},\pmb{w})=1 \ \ \ if \ \ \ \pmb{h} \in
{\mathcal{H}(y_1) \times \ldots \times
\mathcal{H}(y_m)}\\
&P(\pmb{y}|\pmb{h},\pmb{x},\pmb{w})=0 \ \ \ if \ \ \ \pmb{h} \notin
{\mathcal{H}(y_1) \times \ldots \times
\mathcal{H}(y_m)}
\end{split}
\end{equation*}
where $m$ is the length of the labeling\footnote{A labeling is a sequence of predicted classes: $\pmb{y}=y_1,y_2,\ldots,y_m$.}: $m=|\pmb{y}|$.
The formula $\pmb{h} \in
{\mathcal{H}(y_1) \times \ldots \times
\mathcal{H}(y_m)}$ indicates that the latent-labeling $\pmb{h}$ is a latent-labeling of the labeling $\pmb{y}$, which can be more formally defined as follows:
\begin{equation*}
\pmb{h} \in
{\mathcal{H}(y_1) \times \ldots \times
\mathcal{H}(y_m)}
\ \ \ \Longleftrightarrow \ \ \
h_j \!
\in \!\! \mathcal{H}(y_j) \  for \  j=1, \ldots, m.
\end{equation*}
Since
sequences that have any $h_j \notin \mathcal{H}(y_j)$ will by
definition have $P(\pmb{y}|{h}_j,\pmb{x},\pmb{w})=0$,
the model can be simplified as:
\begin{equation} \label{eq.2}
P(\pmb{y}|\pmb{x},\pmb{w})
\triangleq
\sum_{ \pmb{h} \in
{\mathcal{H}(y_1) \times \ldots \times
\mathcal{H}(y_m)}}P(\pmb{h}|\pmb{x},\pmb{w}).
\end{equation}
In Figure~\ref{fig.disjoint}, an example is shown to illustrate the way to compute $P(\pmb{y}|\pmb{x},\pmb{w})$ for a given labeling by summing over probabilities of all its latent-labelings.
The item $P(\pmb{h}|\pmb{x},\pmb{w})$ is defined by the
usual conditional random field formulation:
\begin{equation} \label{eq.3}
 P(\pmb{h}|\pmb{x},\pmb{w})
 =\frac
 {\exp \big\{\pmb{w}^\top \pmb{f}(\pmb{h},\pmb{x})\big\}}
 {\sum\limits_{\pmb{h'} \in \mathcal{H} \times \ldots \times
\mathcal{H}} \exp \big\{\pmb{w}^\top
 \pmb{f}(\pmb{h'},\pmb{x})\big\}},
\end{equation}
in which $\pmb{f}(\pmb{h},\pmb{x})$ is a global feature vector.
LCRF models can be seen as a natural extension of CRF
models, and CRF models can be seen as a special case of LCRFs that
employ only one latent variable for each label (i.e., $|\mathcal{H}(y)| = 1$ for each $y$ in $\mathcal{Y}$).
The global feature vector can be calculated by summing over all its local feature vectors: 
\begin{equation} \label{eq.9}
\pmb{f}(\pmb{h},\pmb{x})=\sum_{i=1,\cdots,m} \pmb{f}_i(\pmb{h},\pmb{x}) + \sum_{i=1,\cdots,m-1} \pmb{f}_{i,i+1}(\pmb{h},\pmb{x}),
\end{equation}
in which $\pmb{f}_i(\pmb{h},\pmb{x})$ represents a local feature vector that depends only on $h_i$ and $\pmb{x}$. The $\pmb{f}_{i,i+1}(\pmb{h},\pmb{x})$ represents the local feature vector that depends only on $h_i$, $h_{i+1}$, and $\pmb{x}$.

Given a training set consisting of $n$ labeled sequences,
$(\pmb{x}_i,\pmb{y}_i)$, for $i=1 \ldots n$, parameter
estimation is performed by optimizing the objective function,
\begin{equation*}
L(\pmb{w})=\sum_{i=1}^n \log
P(\pmb{y}_i|\pmb{x}_i,\pmb{w})-R(\pmb{w}).
\end{equation*}
The first term of this equation represents a conditional marginal
log-likelihood of a training data. The second term is a regularizer
that is used for reducing overfitting in parameter estimation. $L_2$ regularization is often used, and it is defined as:
$R(\pmb{w})=
\frac
{||\pmb{w}||^2}
{2\sigma^2}$, where $\sigma$
is the hyper-parameter that controls the degree of regularization.

\section{Complexity of Exact Inference}
In this section, we will make a formal analysis of the computational complexity of inference in LCRFs. The computational complexity analysis will be based on a reduction from a well-known NP-hard problem to the inference problem on LCRFs.
We assume that the reader has basic background in the complexity theory, including the notions of \emph{NP} and \emph{NP-hardness} \cite{Garey79,Hopcroft79,Lewis81,BartonB87}. Normally, a minimum requirement for an algorithm to be considered as \emph{efficient} is that its running time is \emph{polynomial}: $O(n^c)$, where $c$ is a constant real value and $n$ is the size of the input. It is believed that the NP-hard problems cannot be solved in polynomial time, although there has been no proof of a super-polynomial lower bound.

\begin{figure}[t]
\begin{center}
\begin{tabular}{c}
    \includegraphics[width=0.2\hsize]{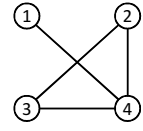}
\end{tabular}
\caption{An example of an undirected graph that contains one maximal clique. This example will be used for complexity analysis.
}\label{fig.max.clique}
\end{center}
\end{figure}

The well-known \emph{\textbf{Maximum-Clique}} problem is an NP-hard problem. Figure~\ref{fig.max.clique} gives an example of the maximum clique problem. The graph consists of 4 connected nodes, and the number of maximum-clique nodes is 3, because the maximum clique is \{2, 3, 4\}.

We will prove that the exact inference in LCRFs is an NP-hard problem, and hence, finding the labeling with the maximum probability on LCRFs is likely to be intractable. Inspired by the consensus string problem on hidden Markov models \cite{Lyngso02}, we establish the hardness analysis of the problem by a reduction from the maximum clique problem.

\begin{definition}
\textbf{Maximum clique problem}\\
\textbf{Instance}: An undirected graph $G=\{\mathcal{V},\mathcal{D}\}$ with the indexed nodes $\mathcal{V}=\{1, 2, \dots, |\mathcal{V}|\}$ and the edge set $\mathcal{D}$ defined among $\mathcal{V}$.\\
\textbf{Question}: What is the size of the maximum clique of $G$?
\end{definition}

Since the $\pmb{x}$ and $\pmb{w}$ are fixed for the inference task, we can simplify the denotation and define \emph{node scores} $\varphi_i(\pmb{h})=\exp \{\pmb{w}^\top \pmb{f}_i(\pmb{h},\pmb{x})\}$ and \emph{edge scores} $\varphi_{i,i+1}(\pmb{h})=\exp \{\pmb{w}^\top \pmb{f}_{i,i+1}(\pmb{h},\pmb{x})\}$.
Then, based on Eq.~\ref{eq.9}, we can reformulate Eq.~\ref{eq.3} as follows:

\begin{equation*}
\begin{split}
 &P(\pmb{h}|\pmb{x},\pmb{w})\\
 &
 =
 \frac
 {\prod\limits_{i=1,\cdots,m} {\varphi_i(\pmb{h})}
 \cdot
 \prod\limits_{i=1,\cdots,m-1} {\varphi_{i,i+1}(\pmb{h})}
 }
 {\sum\limits_{\pmb{h'} \in \mathcal{H} \times \ldots \times
\mathcal{H}}
\Bigg\{ \prod\limits_{i=1,\cdots,m} {\varphi_i(\pmb{h'})}
 \cdot
 \prod\limits_{i=1,\cdots,m-1} {\varphi_{i,i+1}(\pmb{h'})}\Bigg\}}
.
\end{split}
\end{equation*}
This reformulation indicates that the probability of a latent labeling $\pmb{h}$ can be computed by the \emph{path-score} of $\pmb{h}$, and divided by the summation of all path-scores. The \emph{path-score} of  $\pmb{h}$ is defined by
the multiplication over all the node scores and edges scores of $\pmb{h}$.

\begin{definition}
\textbf{Inference in LCRFs}\\
\textbf{Instance}: A latent variable lattice (e.g., see Figure~\ref{fig.disjoint}) $M=\{\pmb{x}, \mathcal{Y}, \mathcal{H}, \Phi_{n}, \Phi_{e}\}$: For input sequence $\pmb{x}$, $|\pmb{x}|=m$; $\mathcal{Y}$ and $\mathcal{H}$ are defined as before;
$\Phi_{n}$ is a $m \times |\mathcal{H}|$ matrix, in which an element $\Phi_{n}(i,j)$ represents a real-value node score $\varphi_n$ for the latent variable $j \in \mathcal{H}$ on the position $i$ (corresponding to $x_i$); $\Phi_{e}$
is a $m \times |\mathcal{H}| \times |\mathcal{H}|$ three-dimensional array, in which an element  $\Phi_{e}(i,j,k)$ represents a real-value edge score $\varphi_e$ for the edge (transition) between the latent variables $j \in \mathcal{H}$ and $k \in \mathcal{H}$, which are on the positions $i$ and $i+1$, respectively.\\
\textbf{Question}: With the $P(\pmb{y}|\pmb{x},\pmb{w})$ being defined in Eq.~\ref{eq.2}, what is the optimal labeling $\pmb{y}^*$ in $M$ such that $\pmb{y}^*=\argmax_{\pmb{y}}
{P(\pmb{y}|\pmb{x},\pmb{w})}$?
\end{definition}

Based on those definitions, we have the following theorem:
\begin{theorem}\label{theo.np.hard}
The computational complexity of the exact inference in LCRFs, $\pmb{y}^*=\argmax_{\pmb{y}}
{P(\pmb{y}|\pmb{x},\pmb{w})}$, is \textbf{NP-hard}.
\end{theorem}
This means the exact inference on LCRFs is an NP-hard problem. The proof is extended from related work on complexity analysis~\cite{ParkD04,Simaan02}. In~\cite{ParkD04}, the complexity analysis is based on Bayesian networks. In ~\cite{Simaan02}, the analysis is based on grammar models. We extend the complexity analysis to LCRFs.

\subsection{Proof}

\begin{figure}[t]
\begin{center}
\begin{tabular}{c}
    \includegraphics[width=0.8\hsize]{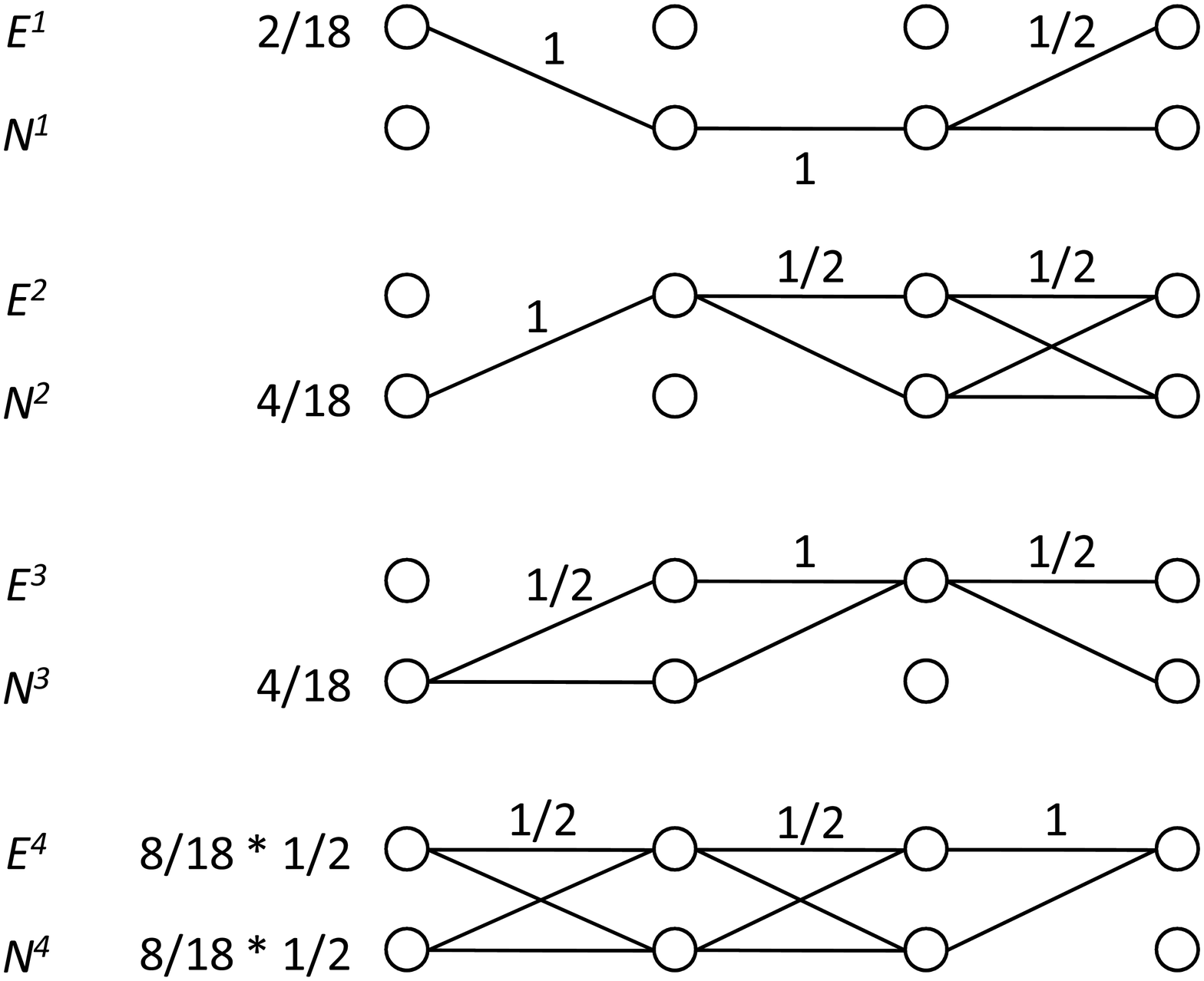}
\end{tabular}
\caption{The latent conditional model $M_G$ constructed from the graph G in Fig~\ref{fig.max.clique}.
}\label{fig.analysis1}
\end{center}
\end{figure}

\begin{figure}[t]
\begin{center}
\begin{tabular}{c}
    \includegraphics[width=0.9\hsize]{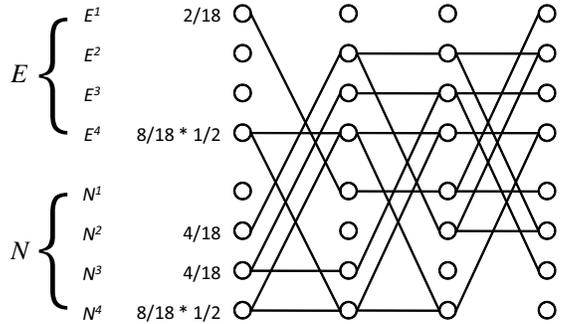}
\end{tabular}
\caption{The same latent conditional model $M_G$ transformed from Fig~\ref{fig.analysis1}. This figure is to show that the latent conditional model constructed in Fig~\ref{fig.analysis1} is valid.
}\label{fig.analysis2}
\end{center}
\end{figure}

Here, we prove the Theorem~\ref{theo.np.hard}.
For an undirected graph $G=\{\mathcal{V},\mathcal{D}\}$, we present the construction of a corresponding latent conditional model $M_G$ from $G$, so that the size of the maximum clique of $G$ is proportional to the probability of the optimal labeling of $M_G$. We set the length of the input sequence: $m=|\mathcal{V}|$. We set $\mathcal{Y}=\{E,N\}$. We set $\mathcal{H}=\{E^1, E^2, \dots, E^{|\mathcal{V}|}, N^1, N^2, \dots, N^{|\mathcal{V}|}\}$, and the disjoint association between $\mathcal{Y}$ and $\mathcal{H}$ is as follows: $\mathcal{H}(E)=\{E^1, E^2, \dots, E^{|\mathcal{V}|}\}$, and $\mathcal{H}(N)= \{N^1, N^2, \dots, N^{|\mathcal{V}|}\}$.
The settings on node-scores and edge-scores on the lattice are as follows:
We group paths with non-zero probability in the lattice into $|\mathcal{V}|$ \emph{layers}, such that each layer contains two horizontal rows of the lattice, and there is a layer corresponding to a node in $\mathcal{V}$. In what follows, we say a path is valid if it has a non-zero probability, and a node is valid if there is at least one valid path that passes the node.
The properties of a layer are as follows:

There are a total of $|\mathcal{V}|$ layers $L=\{L_1, L_2, \dots, L_{|\mathcal{V}|}\}$. A layer $L_i$ corresponds to a node $i \in \mathcal{V}$. Any paths that go through nodes from different layers are not valid. The layer $L_i$ contains only ${E^i}$ and $N^i$. All \emph{paths} (latent-labelings) in the layer should pass the $E^i$ and \emph{avoid} the $N^i$ on the $i$'th position: $\Phi_{n}(i,E^i)=1$ and $\Phi_{n}(i,N^i)=0$.
For any position $k$ other than $i$ in the layer $L_i$, if the node $k$ in $G$ is connected to node $i$ in $G$, $(k,i)\in \mathcal{D}$, then both the ${E^i}$ and $N^i$ are valid on the position $k$: $\Phi_{n}(k,E^i)=1$ and $\Phi_{n}(k,N^i)=1$. Otherwise, only the $N^i$ is valid.
For the edge scores $\Phi_{e}$, all the edges involving an invalid node (with node score of 0) will become an invalid edge (with edge score of 0). For any of the remaining valid edges $(j,k)$, its edge score is 1/2 if node $j$ starts two valid edges and 1 if it starts only one valid edge.
Finally, we adjust the node scores of the beginning nodes. If both of the beginning nodes are valid in a layer, both of the nodes will have the probability of 1/2.
The node scores of the beginning nodes of the layer $L_i$ are multiplied a factor $\delta(i)$ that is related to the degree of the node $i$ in the graph $G$: $\delta_{i}=\frac {2^{deg(i)}} {\sum_{v\in \mathcal{V}}{2^{deg(v)}}}$.

Given any valid path $\pmb{h}$ in the constructed model, it can be proved that the total number of valid paths are $\alpha$, with $\alpha=\sum_{v\in \mathcal{V}}{2^{deg(v)}}$. Also, all the valid paths have the same path-score,
$\frac 1 \alpha$. The summation of all the path scores is exactly 1.
An example for constructing a latent conditional model based on the previous simple graph (see Figure~\ref{fig.max.clique}) is shown in Figure~\ref{fig.analysis1}. In the figure, we only show the valid edges for simplicity.  The path-score of any valid path in Figure~\ref{fig.analysis1} is 1/18.
The model structure in Figure~\ref{fig.analysis1} is a valid type of structure of LCRFs (see Figure~\ref{fig.analysis2}). The reduction is always possible in polynomial time.

\begin{lemma}
If a labeling $\pmb{y}$ in $M_G$ has a probability of $c/\alpha$ ($c$ is a integer with $c \geq 1$), then the graph $G$ must have a maximum clique with the size of at least $c$.
\end{lemma}
Since each valid latent-labeling has the identical probability of $1/\alpha$, $P(\pmb{y}|\pmb{x},\pmb{w})=
c/\alpha$ means that $\pmb{y}$ must have $c$ different latent-labelings.
A clear property of $M_G$ is that one layer can only produce, at most, one latent-labeling for a specified $\pmb{y}$.
Therefore, each of the $c$ latent-labelings of $\pmb{y}$ must come from $c$ different layers. Suppose that the $c$ different layers are $L_{x^1}, L_{x^2}, \dots, L_{x^c}$, and $\mathcal{X}=\{x^1, \dots, x^c\}$ is the set of indexes of the $c$ different layers.
 If $\pmb{y}$ contains a latent-labeling from a layer $L_i$, then $E^i$ must be chosen on the $i$'th position. Therefore, $\pmb{y}$ must have the label $E$ on at least $c$ different positions $x^1, \dots, x^c$. It indicates that each of the $c$ latent-labelings of $\pmb{y}$ must have $E^i$ on at least $c$ different positions $x^1, \dots, x^c$, and for each case, the node $i$ in $G$ is connected to all the $c-1$ nodes indexed by the set $\mathcal{X}-\{i\}$.
 Thus, the nodes $x^1, \dots, x^c$ in $G$ are connected to each other, and they form a clique of the size $c$ in $G$.

\begin{lemma}
If $G$ has a clique of the size $c$,
there must be a labeling in $M_G$ with the probability of at least $c/\alpha$.
\end{lemma}
Suppose that the $c$ nodes of the clique in $G$ are indexed by a set $\mathcal{X}=\{x^1, \dots, x^c\}$; then, in each layer $L_{i}$ with $i \in \mathcal{X}$, there must be a valid $\pmb{h}_{i}$ that passes the $E^{i}$ on all of the positions of $\mathcal{X}$. The $N^{i}$ is always valid for all positions, except the position $i$. Especially, $N^{i}$ is valid for any position $k \notin \mathcal{X}$. On the other hand, $E^i$ is valid at each position $k' \in \mathcal{X}$ since $\mathcal{X}$ forms a clique in $G$. Hence, for $L_{i}$,  the $\pmb{h}_{i}$ described as follows must be valid: $\pmb{h}_{i}$ passes  $E^{i}$ for all positions in $\mathcal{X}$ and passes $N^{i}$ for all positions not in $\mathcal{X}$ (i.e., $\mathcal{V}-\mathcal{X}$). The $c$ latent-labelings from different layers are consistent and belong to an identical labeling, $\pmb{y}$. Since each latent-labeling has the probability of $1/\alpha$, the $\pmb{y}$ has the probability of at least $c/\alpha$.

Combining the two lemmas, we can see that the graph $G$ has a maximum clique of the size $c$ \emph{if and only if} the model $M_G$ has a maximum-probability labeling with the probability $c/\alpha$.
Since the reduction is always possible in polynomial time, we have Theorem~\ref{theo.np.hard}.

\section{A Practical Solution Based on Probability Concentration}
We have shown that exact inference in latent conditional models is an NP-hard
problem. Nevertheless, we try to solve this difficult problem based on an interesting observation. In real world applications, we have an observation on LCRFs: They normally have a highly concentrated probability
distribution. That is, most of the probability mass is distributed on top-$n$ ranked
latent labelings.

\subsection{Probability Concentration from Optimization}
To formally analyze the reason of the probability concentration on LCRFs, we first analyze the optimization process on LCRFs. Since the optimization process on LCRFs is based on the gradient information of its objective function, it is critical to analyze the trends of the gradient formulation of the LCRF objective function. In training LCRFs, people perform gradient ascent for maximizing the objective function. The log-likelihood portion of the objective function is as follows:
\begin{equation*}
\small
\begin{split}
&\mathcal{L}
=
\log \Bigg\{
\frac{\sum_{\pmb{h} \in \pmb{y^*}} \exp{[{\pmb{w}^\top \pmb{f}(\pmb{h},\pmb{x})}]} }
{\sum_{\forall \pmb{h'}} \exp{[{\pmb{w}^\top \pmb{f}(\pmb{h'},\pmb{x})}]} }
\Bigg\}\\
&=
\log \bigg\{
{\sum_{\pmb{h} \in \pmb{y^*}} \exp{[{\pmb{w}^\top \pmb{f}(\pmb{h},\pmb{x})}]} }
\bigg\}
-
\log \bigg\{
{\sum_{\forall \pmb{h'}} \exp{[{\pmb{w}^\top \pmb{f}(\pmb{h'},\pmb{x})}]} }
\bigg\}.
\end{split}
\end{equation*}
Hence, its gradient is as follows:
\begin{equation}\label{eq.5}
\footnotesize
\begin{split}
&\nabla_{\pmb{w}} \mathcal{L}
=
\sum_{\pmb{h} \in \pmb{y^*}}
\Big\{
P^*(\pmb{h})\pmb{f}(\pmb{h},\pmb{x})
\Big\}
-
\sum_{\forall \pmb{h'}}
\Big\{
P(\pmb{h'})\pmb{f}(\pmb{h'},\pmb{x})
\Big\}\\
&=
\sum_{\pmb{h} \in \pmb{y^*}}
\Big\{
[P^*(\pmb{h})- P(\pmb{h})]\pmb{f}(\pmb{h},\pmb{x})
\Big\}
-
\sum_{\pmb{h'} \notin \pmb{y^*}}
\Big\{
P(\pmb{h'})\pmb{f}(\pmb{h'},\pmb{x})
\Big\}
,
\end{split}
\end{equation}
where $P^*(\pmb{h})$ is the probability of $\pmb{h}$ with regard to only $\pmb{y^*}$. In other words,
$$
P^*(\pmb{h})=
\frac
{
\exp[{\pmb{w}^\top \pmb{f}(\pmb{h},\pmb{x})}]
}
{\sum_{\pmb{h} \in \pmb{y^*}} \exp[{\pmb{w}^\top \pmb{f}(\pmb{h},\pmb{x})}] }.
$$
From Equation~\ref{eq.5}, we can see that a latent labeling $\pmb{h} \in \pmb{y^*}$ with higher probability can ``dominate'' the gradient with higher degree. Since the LCRF model is trained with gradient ascent,  a latent labeling $\pmb{h} \in \pmb{y^*}$ with higher probability will in turn be updated with more gains in the next gradient ascent step (because it ``dominated'' the current gradient with more degrees). Hence, we expect latent labelings will have a ``rich get richer'' trend during the training of a LCRF model. This ``rich get richer'' trend is also meaningful in real-world data and tasks, because in this way the latent structure can be discovered with higher confidence.

On the other hand, note that the probability is expected to be concentrated \emph{highly}, but not \emph{completely}. This is because real-world tasks are usually ambiguous. Another reason is from regularization terms of the objective function, which controls overfitting, including the potential overfitting of latent structures.

\subsection{Latent-Dynamic Inference}
Based on the highly (but not completely) concentrated probabilities in LCRFs, we propose a novel inference method, which is efficient and exact in most of the real-world applications.

\subsection{Latent-Dynamic Inference (LDI-Naive)}
In the inference stage, given a test sequence $\pmb{x}$, we want
to find the most probable label sequence, $\pmb{y}^*$:
\begin{equation}
\pmb{y}^*=\argmax_{\pmb{y}}
{P(\pmb{y}|\pmb{x},\pmb{w})}.
\end{equation}
For latent conditional models like LCRFs, the $\pmb{y}^*$ cannot
directly be produced by the Viterbi algorithm, because of the
incorporation of latent variables.

\begin{figure}[t]
\begin{center}
\begin{tabular}{c}\hline
\begin{minipage}{1\hsize}
\begin{algorithmic}[1]

\State {\bf Definitions:}\\
{``$n$'' represents the current search step (\# of latent-labelings being searched).\\
``ProbGap'' is a real value recording the difference between $P(\pmb{y}')$ and $P_{remain}$.\\
``$\mathcal{S}$'' indicates a set of ``already searched labelings''.\\
``$\mathrm{FindLatentLabeling}(n)$'' uses $A^*$ search to find the $n$'th ranked latent-labeling.\\
$``\mathrm{FindParentLabeling}(\pmb{h})$'' finds the corresponding labeling from the latent-labeling:  $\mathrm{FindParentLabeling}(\pmb{h})=\pmb{y} \Longleftrightarrow h_j \in
\mathcal{H}(y_j) \  for \  j=1 \ldots m$.\\
    $P(\pmb{h})\triangleq P(\pmb{h}|\pmb{x},\pmb{w})$.\\
    $P(\pmb{y})\triangleq P(\pmb{y}|\pmb{x},\pmb{w})$.}
    \State {}

\State {\bf Initialization:}\\
    {\it $n=0$; $\mathrm{ProbGap}=-1$;
 $P(\pmb{y}')=0$; $\mathcal{S}_0=\emptyset$.}
 \State {}
\State {{\bf Procedure} LDI():}
   \While {$\mathrm{ProbGap}<0$}
        \State $n=n+1;$
        \State $\pmb{h}_n=\mathrm{FindLatentLabeling}(n);$
        \State $\pmb{y}_n=\mathrm{FindParentLabeling}(\pmb{h}_n);$
        \If{$\pmb{y}_n \notin \mathcal{S}_{n-1}$}

            \State $\mathcal{S}_{n}=\mathcal{S}_{n-1} \cup \{\pmb{y}_n\};$
            \State $P(\pmb{y}_n)=\mathrm{ComputeProbability}(\pmb{y}_n);$

            \If{$P(\pmb{y}_n)>P(\pmb{y}')$}
                \State $\pmb{y}'=\pmb{y}_n;$
            \EndIf

            \State $P_{remain}=1- \sum_{\pmb{y}_k \in \mathcal{S}_n}
    P(\pmb{y}_k);$

            \State $\mathrm{ProbGap}=P(\pmb{y}') -
              P_{remain};$
        \Else
            \State $\mathcal{S}_{n}=\mathcal{S}_{n-1};$
        \EndIf
    \EndWhile
    \State {{\bf return} $\pmb{y}'$;}

\State {}
\end{algorithmic}
\end{minipage}
\\ \hline
\end{tabular}
\caption{The algorithm of the LDI inference for LCRFs.} \label{fig.algo.LDI}
\end{center}
\end{figure}

In this section, we describe an exact inference algorithm, the
latent-dynamic inference (LDI), for producing the optimal label
sequence $\pmb{y}^*$ on LCRFs (see Figure~\ref{fig.algo.LDI}).
In short, the algorithm generates the best latent-labelings in the order
of their probabilities. Then, the algorithm maps each of these latent-labelings to its associated
labelings and uses a dynamic programming method (the forward-backward algorithm) to compute the probabilities of the corresponding labelings.
The algorithm continues to generate the next best latent-labeling and the
associated labeling until there is not enough probability mass
left to beat the best labeling.

In detail, an $A^*$ search algorithm \cite{HartN68,Russell02} with a
Viterbi heuristic function is adopted to produce top-$n$ latent-labelings,
$\pmb{h}_1, \pmb{h}_2, \ldots \pmb{h}_n$. In addition, a
forward-backward-style algorithm is used to compute the
probabilities of their corresponding labelings, $\pmb{y}_1,
\pmb{y}_2, \ldots \pmb{y}_n$. The algorithm then tries to
determine the optimal labeling based on the top-$n$ statistics,
without enumerating the remaining low-probability labelings, in which the number is exponentially large.

The optimal labeling $\pmb{y}^*$ will be $\pmb{y}'$ when the following
``exact-condition'' is achieved:
\begin{equation} \label{eq.6}
P(\pmb{y}'|\pmb{x},\pmb{w})-\Big(1- \sum_{\pmb{y}_k
\in \mathcal{S}_n} P(\pmb{y}_k|\pmb{x},\pmb{w})\Big) \geq
0,
\end{equation}
where $\pmb{y}'$ is the most probable label sequence in the current
stage. It is straightforward to prove that
$\pmb{y}^*=\pmb{y}'$, and further search is unnecessary. This
is because the remaining probability mass, $1- \sum_{\pmb{y}_k
\in \mathcal{S}_n} P(\pmb{y}_k|\pmb{x},\pmb{w})$,
cannot beat the current optimal labeling in this case.

\begin{theorem}\label{theo.2}
In the procedure defined in Figure~\ref{fig.algo.LDI}, the labeling $\pmb{y}'$ (satisfying the exact condition Eq.~\ref{eq.6}) is guaranteed to be the exact optimal labeling:
\begin{equation*}
\pmb{y}'=\argmax_{\pmb{y}}
{P(\pmb{y}|\pmb{x},\pmb{w})}.
\end{equation*}
\end{theorem}
The proof of Theorem~\ref{theo.2} is simple.
Given the \emph{exact condition},
suppose there is a label sequence $\pmb{y}''$ with a larger
probability,
$
P(\pmb{y}''|\pmb{x},\pmb{w}) >
P(\pmb{y}'|\pmb{x},\pmb{w}),
$
then it follows that $\pmb{y}'' \notin \mathcal{S}_n$.
Since $P(\pmb{y}''|\pmb{x},\pmb{w}) >
P(\pmb{y}'|\pmb{x},\pmb{w})$ and $\pmb{y}'' \notin \mathcal{S}_n$, it follows that
$
 P(\pmb{y}''|\pmb{x},\pmb{w})
 + \sum_{\pmb{y}_k \in \mathcal{S}_n}
 P(\pmb{y}_k|\pmb{x},\pmb{w})
 >
 P(\pmb{y}'|\pmb{x},\pmb{w})
 + \sum_{\pmb{y}_k \in \mathcal{S}_n}
 P(\pmb{y}_k|\pmb{x},\pmb{w})
 .
$
According to the exact condition, it follows that
$
 P(\pmb{y}'|\pmb{x},\pmb{w})
 + \sum_{\pmb{y}_k \in \mathcal{S}_n}
 P(\pmb{y}_k|\pmb{x},\pmb{w})
 \geq
 \Big(1- \sum_{\pmb{y}_k
 \in \mathcal{S}_n} P(\pmb{y}_k|\pmb{x},\pmb{w})\Big)
 +\sum_{\pmb{y}_k
 \in \mathcal{S}_n} P(\pmb{y}_k|\pmb{x},\pmb{w}).
$
The right side of the inequality is 1.
Therefore, we have
$
 P(\pmb{y}''|\pmb{x},\pmb{w})
 + \sum_{\pmb{y}_k \in \mathcal{S}_n}
 P(\pmb{y}_k|\pmb{x},\pmb{w})
 >1
 ,
$
which is not possible. Hence, the assumed $\pmb{y}''$ is impossible.

\subsection{Admissible and Tight Heuristics for Efficient Search}

We have presented the framework of the LDI inference. Here, we
describe the details on implementing its crucial component:
designing the heuristic function for the $A^*$ heuristic search.
Our heuristic search aims at finding the top-$n$ most probable latent-labelings. Recall that the probability of a latent-labelings is defined as
$$
P(\pmb{h}|\pmb{x},\pmb{w})
 =\frac
 {\exp \big\{\pmb{w}^\top \pmb{f}(\pmb{h},\pmb{x})\big\}}
 {\sum\limits_{\pmb{h'} \in \mathcal{H} \times \ldots \times
\mathcal{H}} \exp \big\{\pmb{w}^\top
 \pmb{f}(\pmb{h'},\pmb{x})\big\}}
.
$$
To find out top-$n$ most probable latent-labelings, an easier way for achieving the same target is to find out top-$n$ ``highest-score'' latent-labelings with the score defined as
$$
 \varphi(\pmb{h}|\pmb{x},\pmb{w})
 =
\pmb{w}^\top \pmb{f}(\pmb{h},\pmb{x})
.
$$
In $A^*$ search, the cost function is normally defined as:
\begin{equation*}
f(i) = g(i) + heu(i),
\end{equation*}
where $i$ is the current node.
    $f(i)$ is the cost function.
    $g(i)$ is the cost from the start node to the current node.
    $heu(i)$ is the estimated cost from current node to the target node. If the heuristic function is \emph{non-admissible}, the $A^*$ algorithm may overlook the optimal solution \cite{Russell02}. In our case, a heuristic is \emph{admissible} if it never underestimates the scores from the current position to the target.
Note that, admissible heuristics do not guarantee the efficiency of the search. In our LDI case, we not only want admissible heuristics, but also try to make the search efficient enough. For this concern, we need a \emph{monotone} (or, \emph{consistent}) heuristic. A monotone heuristic is an admissible heuristic with additional properties. Informally, we can think a monotone heuristic is an \emph{admissible and tight} heuristic. A formal definition of monotone heuristics is as follows for the highest-score path search problem:
\begin{equation*}
\begin{split}
&
heu(j) \geq c(j,k) + heu(k), and
\\
&
heu(G)=0,
\end{split}
\end{equation*}
where $j$ is every possible current node, and $k$ is every possible successor of $j$ generated by any possible action $a(j,k)$ with the cost $c(j,k)$. $G$ is the goal node.
Here we present a monotone heuristic function for the LDI task. The LDI algorithm first scans in a backward direction (right to left) to compute the monotone heuristic function for each latent variables in the lattice. After that, the $A^*$ search was performed in a forward direction (left to right) based on the computed heuristics. For a latent variable $h^j$ on the position $i$, its monotone heuristic function is designed as follows:
$
heu(i,j)
=\max_{\pmb{h}^{'}
\in \mathcal{L}(i,j)}
\varphi(\pmb{h}^{'}|\pmb{x},\pmb{w})
,
$
where $\mathcal{L}(i,j)$ represents a set of all possible partial
latent-labelings starting from the latent variable $h^j \in \mathcal{H}$ on position $i$ and ending at the goal position $m$.
In implementation, the heuristics are computed efficiently by using a Viterbi-style algorithm:
\begin{equation*}
\begin{split}
&
(1)\ \ Initialization:\\
&
for \ \ j=1,\dots,|\mathcal{H}|, \ \
heu(m,j)
=0.
\\
&
(2)\ \ Recursion \ \ (for \ \ i=m-1,\dots,1):\\
&
for \ \ j=1,\dots,|\mathcal{H}|,\\
&heu(i,j)
=\max_{k=1,\dots,|\mathcal{H}|}
[heu(i+1,k)+\Phi_{n}(i+1,k)+\Phi_{e}(i,j,k)].
\end{split}
\end{equation*}
The $\Phi_{n}(i+1,k)$ is the score of the latent variable $h^k$ on the position $i+1$; the $\Phi_{e}(i,j,k)$ is the score of the edge between the latent variable $h^j$ on the position $i$ and the following latent variable $h^k$ on the next position.

\subsection{A Bounded Version (LDI-Bounded)}

\begin{figure}[t]
\begin{center}
\begin{tabular}{c}\hline
\begin{minipage}{1\hsize}
\begin{algorithmic}[1]

\State {{\bf Procedure} LDI-Bounded($n'$):}
   \While {$\mathrm{ProbGap}<0$}
        \State $n=n+1;$
        \If{$n>n'$}
                \State {{\bf return} $\pmb{y}'$;}
        \EndIf

        \State $\pmb{h}_n=\mathrm{FindLatentLabeling}(n);$
        \State {...}
    \EndWhile
    \State {{\bf return} $\pmb{y}'$;}
    \State {}

\end{algorithmic}
\end{minipage}
\\ \hline
\end{tabular}
\caption{The algorithm of the LDI-Bounded inference for LCRFs. Since a majority of the steps are similar to the Figure~\ref{fig.algo.LDI}, we do not repeat the description. The new input variable $n'$ ($n'\geq 1$) represents the threshold value for bounding the search steps.} \label{fig.algo.LDI.bound}
\end{center}
\end{figure}

By simply setting a threshold value on the search step, $n$, we can
derive a bounded version of the LDI; i.e., LDI-Bounded (see Figure~\ref{fig.algo.LDI.bound}). This method is a
straightforward way for approximating the LDI. We
have also tried other methods for approximation. Intuitively, one
alternative method is to design an approximated ``exact condition",
by using a factor, $\alpha$, to estimate the distribution of the
remaining probability:

\begin{equation*}
P(\pmb{y}'|\pmb{x},\pmb{w})-\alpha \Big(1-
\!\!\sum_{\pmb{y}_k \in \mathcal{S}_n}\!\!
P(\pmb{y}_k|\pmb{x},\pmb{w})\Big) \geq 0.
\end{equation*}
For example, if at most 50\% of the unknown
probability, $1- \sum_{\pmb{y}_k \in \mathcal{S}_n}
P(\pmb{y}_k|\pmb{x},\pmb{w})$, can be distributed on a
single labeling, we can set $\alpha=0.5$ to make a loose condition
to stop the inference. At first glance, this seems to be quite
intuitive. However, when we compared this alternative method with the
LDI-Bounded method, we found that the performance and speed of the former method was worse than for the latter.

\subsection{Existing Inference Methods on Latent Conditional Models}
In \cite{MatsuzakiM05}, the optimal labeling is approximated by using a modified Viterbi inference (MVI) method. In the MVI inference, there are two steps. First, the MVI searches for the optimal latent-labeling using the Viterbi algorithm:
\begin{equation*}
\pmb{h}^*=\argmax_{\pmb{h}}
{P(\pmb{h}|\pmb{x},\pmb{w})}.
\end{equation*}
Then, a labeling $\pmb{y}$ is derived by directly locating the corresponding labeling of the latent-labeling $\pmb{h}^*$:
\begin{equation*}
\pmb{y}=\mathrm{FindParentLabeling}(\pmb{h}^*),
\end{equation*}
which means that $h_j \in
\mathcal{H}(y_j)$ for $j=1 \ldots m$.
The MVI inference can be seen as a
simple adaptation of the traditional Viterbi inference in the case of latent conditional models.

In \cite{MorencyQ07}, $\pmb{y}^*$ is
estimated by a point-wise marginal inference (PMI) method. To
estimate the label ${y}_j$ of token $x_j$, the marginal probabilities
$P(h_j|\pmb{x},\pmb{w})$ are computed for all
possible latent variables $h_j \in \mathcal{H}$. Then the marginal
probabilities are summed up (according to the association between latent variables and labels) for computing $P(y_j|\pmb{x},\pmb{w})$ for all possible labels $y_j \in \mathcal{Y}$. In this way, the optimal labeling is approximated by choosing the labels with the maximum marginal probabilities at each position $j$ independently:
\begin{equation*}
\pmb{y}=\argmax_{y_j \in \mathcal{Y}}
{P(y_j|\pmb{x},\pmb{w})} \ \ for \ \ j=1,\dots,m,
\end{equation*}
where
\begin{equation*}
P(y_j|\pmb{x},\pmb{w})=\frac {\sum_{\pmb{h} \in \mathcal{H}(y_j)} P(\pmb{h}|\pmb{x},\pmb{w})}
{\sum_{\pmb{h} \in \mathcal{H}} P(\pmb{h}|\pmb{x},\pmb{w})}.
\end{equation*}

The LDI-Naive and the LDI-Bounded perform exact inference or almost-exact inference, while the MVI and the
PMI perform a rough estimation on $\pmb{y}^*$. We will compare the different methods via
experiments in Section \ref{sec:experiments}.

\subsection{Comparison with MAP Algorithms}
The MAP problem refers to finding the Maximum a Posteriori hypothesis, which aims at finding the most likely configuration of a set of variables in a Bayesian network, given some partial evidence about the complement of that set. Several researchers have proposed algorithms for solving the MAP problem \cite{ParkD03,ParkD04,Yanover03,SunDY07}.

In \cite{ParkD04}, an efficient approximate local search algorithm is proposed for approximating MAP: \emph{hill climbing} and \emph{taboo search}.
Compared to the approximate local search algorithm, the LDI algorithm can perform exact inference under a reasonable number of search steps (with a tractable cost). In the case that exactitude is required, this characteristic of the LDI algorithm is important.
In \cite{SunDY07}, a dynamic weighting $A^*$ ($DWA^*$) search algorithm is proposed for solving MAP in Bayesian networks. Like the local search algorithms,
the $DWA^*$  search is an approximate method and it does not guarantee an exact solution.
In \cite{ParkD03}, an effective method is proposed to compute
a relatively tight upper-bound on the probability of a MAP solution. The upper
bound is then used to develop a branch-and-bound search algorithm for solving MAP exactly.
Whether or not the branch-and-bound search can be used for solving LCRFs is unclear, because of the structural difference. In addition, the quality of tightness of the computed bound is crucial for the tractability of the branch-and-bound search. The quality of tightness is unclear concerning LCRFs.

\section{Conclusions}
We made a formal analysis of the inference in latent conditional models, and showed that it is an NP-hard problem, even when latent conditional models have a disjoint assumption and linear-chain structures.
More importantly, based on an observation of probability concentration, we proposed the latent-dynamic inference method (LDI-Naive) and its bounded version (LDI-Bounded), which are able to
perform exact and fast inference in latent conditional models, even though the original problem is NP-hard.



\end{document}